\def\year{2022}\relax
\documentclass[letterpaper]{article} 
\usepackage{aaai22}  
\usepackage{times}  
\usepackage{helvet}  
\usepackage{courier}  
\usepackage[hyphens]{url}  
\usepackage{graphicx} 
\usepackage{amsmath}
\usepackage{multirow}
\usepackage{xcolor}
\usepackage[toc,page]{appendix}

\usepackage{longtable}
\usepackage{natbib}  
\usepackage{caption} 
\DeclareCaptionStyle{ruled}{labelfont=normalfont,labelsep=colon,strut=off} 
\usepackage{algorithm}
\usepackage{algorithmic}

\usepackage{newfloat}
\usepackage{listings}
\floatstyle{ruled}
\newfloat{listing}{tb}{lst}{}
\floatname{listing}{Listing}

\newcommand{\notshortstack}[2][l]{%
  \begin{tabular}{@{}#1@{}}#2\end{tabular}%
}

\title{SGD-X: A Benchmark for Robust Generalization in Schema-Guided Dialogue Systems}
\author {
    Harrison Lee\footnote{Equal contribution},
    Raghav Gupta\footnotemark[1],
    Abhinav Rastogi,
    Yuan Cao,
    Bin Zhang,
    Yonghui Wu
}
\affiliations {
    Google Research\\
    \{harrisonlee,raghavgupta,abhirast,yuancao,zbin,yonghui\}@google.com
}

\begin{document}

\maketitle

\begin{abstract}
Zero/few-shot transfer to unseen services is a critical challenge in task-oriented dialogue research. The Schema-Guided Dialogue (SGD) dataset introduced a paradigm for enabling models to support any service in zero-shot through \textit{schemas}, which describe service APIs to models in natural language. We explore the robustness of dialogue systems to linguistic variations in schemas by designing SGD-X - a benchmark extending SGD with semantically similar yet stylistically diverse variants for every schema. We observe that two top state tracking models fail to generalize well across schema variants, measured by joint goal accuracy and a novel metric for measuring schema sensitivity. Additionally, we present a simple model-agnostic data augmentation method to improve schema robustness.
\end{abstract}

\renewcommand{\arraystretch}{0.91}

\section{Introduction}
\label{sec:intro}
\noindent Task-oriented dialogue systems have begun changing how we interact with technology, from personal assistants to customer support. One obstacle preventing their ubiquity is the resources and expertise needed for their development. Traditional approaches operate on a fixed ontology \cite{henderson2014word, mrkvsic2017neural}, which is not suited for a dynamic environment. For every new service that arises or modification to an existing service, training data must be re-collected and systems re-trained.

\begin{figure*}[ht]
\centering
\includegraphics[width=0.95\textwidth]{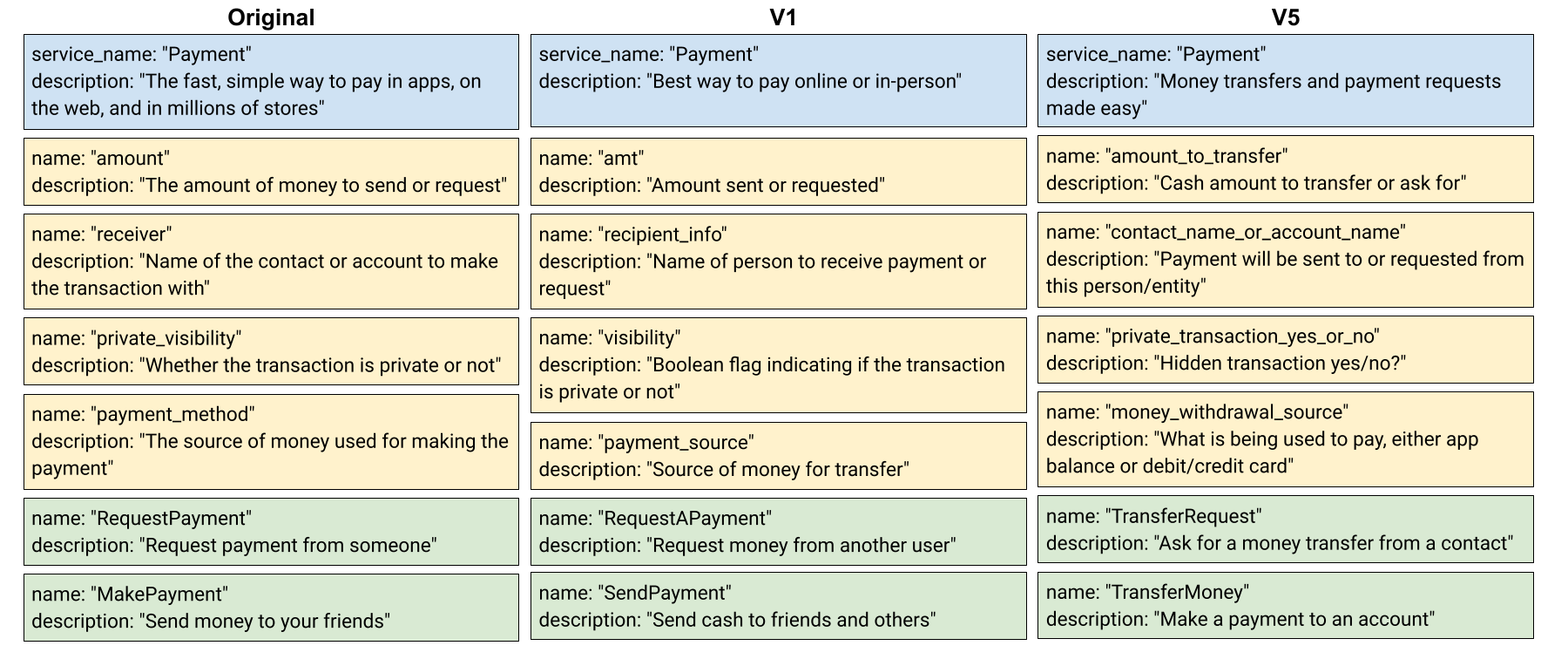}
\caption{The original schema for a Payment service (left) alongside its closest and farthest SGD-X variants (center and right, respectively), as measured by linguistic distance functions. We study the robustness of models to writing styles used in schemas.}
\label{fig:schema-variations}
\end{figure*}

The Schema-Guided Dialogue paradigm, introduced in \citet{rastogi2020towards}, advocates for the creation of a universal dialogue system which can interface with any service, without service or domain-specific optimization. Each service is represented by a \textit{schema}, which enumerates the slots and intents of the service and describes their functionality in natural language (see Figure \ref{fig:schema-variations}). Schema-guided systems interpret conversations, execute API calls, and respond to users based on the schemas provided to it. In theory, this enables a single system to support any service; in practice, whether this is feasible hinges on how robustly models can generalize beyond services seen during training.

In the Schema-Guided Dialogue challenge at DSTC8 \citep{rastogi2020schema}, participants developed schema-guided dialogue state tracking models, which were evaluated on both seen and unseen services. While results were promising, with the top team achieving 87\% \textit{joint goal accuracy} (92\% seen, 85\% unseen), we observed a major shortcoming with SGD - the dataset's schemas are unrealistically uniform compared to the diverse writing styles encountered ``in the wild", where schemas are written by API developers of various backgrounds. 

The uniformity of SGD is evident in its schema element names. Of the names in the test set schemas ``unseen" in the train set, 71\% of intent names and 65\% of slot names exactly match names appearing in the train schemas, meaning most names in ``unseen" schemas are actually already seen by the model during training. MultiWOZ \citep{budzianowski2018multiwoz}, another popular dialogue state tracking benchmark, faces similar issues in the zero-shot leave-one-domain-out setup \citep{wu2019transferable}, with 60-100\% of slot names in the held-out domain seen by the model during training. SGD descriptions are also uniformly written. For example, all descriptions for boolean slots either begin with the phrase ``Boolean flag..." or ``Whether...".

We hypothesize that the uniformity of SGD schemas allows models to overfit on specific linguistic styles without penalty in evaluation, leading to an overestimate of the generalizability of models. Additionally, ``seen" schemas in evaluation are identical to the ones seen in training, meaning SGD does not evaluate how well models handle changes in seen schemas, however minor.

In this work, we investigate the robustness of schema-guided models to linguistic styles of schemas. Our contributions are as follows:

\begin{itemize}
\item We introduce SGD-X, an extension to the SGD dataset that contains crowdsourced stylistic variants for every schema in the original dataset\footnote{We release SGD-X and an evaluation script for schema-guided dialogue state tracking models on GitHub at https://github.com/google-research-datasets/dstc8-schema-guided-dialogue} 
\item Based on SGD-X, we propose \textit {schema sensitivity} - a metric to evaluate model sensitivity to schema variations
\item We show that two top schema-guided dialogue state tracking (DST) models based on BERT and T5 are highly sensitive to schema variations, dropping 12-18\% in joint goal accuracy for the average SGD-X variant
\item We demonstrate that back-translation is an effective, model-agnostic technique for improving schema robustness
\end{itemize}

\section{The SGD-X Dataset}
We curate SGD-X, short for \textit{Schema Guided Dialogue - eXtended}, to evaluate the robustness of schema-guided dialogue models to schemas. Following SGD terminology, we define a \textit{schema} as a collection of intents and slots belonging to a service, along with metadata that describe their intended behavior. We also define a \textit{schema element} as an intent, slot, or service identifier. A key feature of schemas is the inclusion of natural language descriptions for each schema element. For example, an intent ``\textit{SearchMap}" might have the description ``\textit{Search for a location of interest on the map}".

For every schema in SGD, SGD-X provides 5 variants, where each one replaces the original schema element names and descriptions with semantically similar paraphrases. Figure \ref{fig:schema-variations} shows an original schema alongside two SGD-X variants. We describe the dataset in detail below.

\subsection{Blind Paraphrase Collection}

Schema element names and descriptions in the original SGD dataset were written by a small set of authors, and achieving linguistic diversity was not an explicit goal. To diversify SGD-X, we crowdsourced paraphrases across 400+ authors from Amazon Mechanical Turk. We chose crowdsourcing over automatic paraphrasing methods because we found that automatic methods were often semantically inaccurate and provided insufficient linguistic diversity, especially when the text was short. We designed two crowdsourcing tasks (pictured in Figure \ref{fig:endor-intent-name} and \ref{fig:endor-slot-desc}):

\textbf{Paraphrasing names:} 
To paraphrase names, we provided a schema element's long-form description from the SGD dataset and asked crowdworkers to generate a short name that would capture the description. We deliberately did not share the original names to encourage a diversity of paraphrases - hence ``blind" paraphrasing.

\textbf{Paraphrasing descriptions:} To generate descriptions, we reversed the name paraphrasing task - i.e. given only the name of a schema element, we asked crowdworkers to come up with a long-form description. For a limited set of schema elements, we provided additional information:
\begin{itemize}
    \item If intent and slot names were ambiguous on their own (e.g. the ``\textit{intent}" slot from the Homes service, which indicates whether a user is interested in buying or renting property), the original description was shown
    \item For categorical slots, their possible values were shown
\end{itemize}

For a single task, a crowdworker was tasked to come up with either all names or all descriptions for a given service's schema elements.

After collecting raw responses, we deduplicated and manually vetted responses for quality and correctness. Our primary criterion was whether a response accurately described the schema element, and sometimes valid responses did not fully overlap semantically with the original as traditional paraphrasing typically requires. For instance, we considered \texttt{SearchByLocation} a valid replacement for \texttt{Find\textbf{Home}ByArea}, despite the former's lack of reference to the ``home" concept, since it is implied that the search is for homes in the broader context of the \texttt{Homes} service.

We created enough tasks to collect approximately 10 paraphrases per schema element name and description. At the end of the collection and vetting phase, we had at least 5 paraphrases for every name and description. When there were more than 5, we selected 5 at random.

\subsection{Composing Schema Variants}
We composed our schema element paraphrases into schema variants, where each variant replaces every name and description in the original schema with a crowdsourced paraphrase. We placed paraphrases into schema variants such that variants increasingly diverge from the original schemas as the variant number increases. We sorted each schema element's name/description paraphrases by their distance from the original name/description using the following metrics:
\begin{itemize}
    \item For names, we used Levenshtein distance
    \item For descriptions, we used Jaccard distance, where stopwords were removed and words were lemmatized using spaCy \citep{spacy}
\end{itemize}

After sorting, for every schema element $elem$, we obtained a list of unique name paraphrases $N^{elem} = [n^{elem}_{idx}], idx\in\{1..5\}$, ordered by increasing Levenshtein distance from the original name $n^{elem}_{gt}$. Similarly for every schema element description, we obtained a list of unique description paraphrases $D^{elem} = [d^{elem}_{idx}], idx\in\{1..5\}$, ordered by increasing Jaccard distance from the original description $d^{elem}_{gt}$.

Finally to compose the $idx$ schema variant, for every $elem$ in the schema, we simply selected $n^{elem}_{idx}$ and $d^{elem}_{idx}$. This establishes the SGD-X benchmark as a series of increasingly challenging evaluation sets. Henceforth in this paper, we refer to these schema variants as $v_1$ through $v_5$, where $v_1$ refers to the variant schema closest to the original and $v_5$ the farthest. Figure \ref{fig:schema-variations} compares an original schema with its first and fifth variant to highlight the increasing divergence property.

\subsection{Dataset Statistics}

The original SGD dataset contains 45 schemas with a total of 365 slots and 88 intents. Each schema element is associated with 1 name and 1 description (though service names were not paraphrased). After compiling paraphrases into variant schemas, SGD-X presents 5 variants for every schema, totalling 4,755 paraphrases. Each schema variant is composed of paraphrases from multiple crowdworkers. Designing the tasks, collecting data, manually vetting responses, and composing the variants took approximately 1 month. 

Table \ref{tab:sgdx-stats} presents various metrics on SGD-X. As mentioned in Section \ref{sec:intro}, one concern with the original test set is that roughly 70\% of the slot and intent names in the 15 ``unseen" schemas appear in training schemas. In contrast, that figure drops to 8\% for slot names and 2\% for intent names for the average SGD-X variant.

For names, the average normalized Levenshtein distance from original to paraphrase is about 0.5 (on a scale of 0 to 1), indicating high variation. For descriptions, the average BLEU score between original and paraphrase is 7.9, and the average BLEU score among paraphrased descriptions (i.e. self-BLEU\footnote{We calculate self-BLEU for a description by calculating the BLEU score between every pair of variants, resulting in 5 * 4 = 20 scores. We then compute top-line self-BLEU by averaging these scores across all descriptions across all 45 unique schemas.}) is 4.5, indicating a large diversity of descriptions.

\begin{table}[]
\large
\centering
\setlength{\tabcolsep}{3pt}
\scalebox{0.87}{
\begin{tabular}{|l|c|ccccc|c|}
\hline & &\multicolumn{6}{c|}{Schema variant}\\\hline
Metric & Orig & v1 & v2 & v3 & v4 & v5 & \textbf{Avg} \\\hline
\hline
{\notshortstack{\% of test slot\\names seen in train}} & 65\% & 13\% & 14\% & 5\% & 6\% & 2\% & \textbf{8\%}\\\hline
{\notshortstack{\% of test intent\\names seen in train}} & 71\% & 0\% & 0\% & 4\% & 0\% & 4\% & \textbf{2\%}\\\hline
{\notshortstack{Levenshtein\\ Distance (names)}} & - & 0.30 & 0.42 & 0.49 & 0.56 & 0.61 & \textbf{0.48}\\\hline
BLEU (desc) & - & 18.8 & 11.3 & 5.6 & 2.9 & 1.0 & \textbf{7.9}\\\hline
\end{tabular}}
\caption{SGD-X dataset statistics. The metrics show high linguistic variation from the original SGD schemas.}
\label{tab:sgdx-stats}
\end{table}

\section{Evaluation Methodology}

We propose evaluating models by training them on original SGD only and evaluating on SGD-X. In addition to standard accuracy metrics, we propose measuring the consistency of predictions across variants. Below, we first describe our schema sensitivity metric, followed by a general proposal for training and evaluating dialogue systems on SGD-X, and finally a detailed proposal for evaluating dialogue state tracking models specifically.

\subsection{Schema Sensitivity Metric}

Let $\mathcal{M}$ be a turn-level evaluation metric, which takes a prediction and ground truth at turn $t$ as input and returns a score. Let $K$ denote the number of schema variants, $p^k_t$ denote turn-level predictions for variant $k$, and $g_t$ denote the ground-truth. We define \textit{schema sensitivity} ($SS$) for the metric $\mathcal{M}$ as the turn-level Coefficient of Variation ($CoV$) of the metric value (i.e., the standard deviation normalized by the mean) averaged over all turns in the evaluation set. This is described by the following set of equations:

\begin{equation} \label{eq1}
SS_\mathcal{M} = \frac{1}{|T|}\sum\limits_{t \in T} CoV_t = \frac{1}{|T|}\sum\limits_{t \in T} \frac{s_t}{\bar{x}_t}
\end{equation}

where the standard deviation $s_t$ and mean $\bar{x}_t$ are defined as follows:

\begin{equation} \label{eq2}
s_t = \sqrt{\frac{\sum\limits_{k=1}^{K} (\mathcal{M}(p^k_t, g_t) - \overline{\mathcal{M}}(\mathbf{p_t}, g_t))^2}{K - 1}}
\end{equation}

\begin{equation} \label{eq3}
\bar{x}_t = \overline{\mathcal{M}}(\mathbf{p_t}, g_t)
\end{equation}

$\overline{\mathcal{M}}(\mathbf{p_t}, g_t) = \frac{1}{K}\sum\limits_{k=1}^K \mathcal{M}(p_t^k, g_t)$ is the average of the metric corresponding to predictions over all $K$ variants in turn $t$, and $T$ is the set of all turns in the eval set. 

Intuitively, schema sensitivity quantifies how much predictions fluctuate when exposed to schema variants, independent of the prediction correctness, and models with lower $SS$ are more robust to schema changes. $SS$ may be computed for any turn-level or dialogue-level metric across the schema-guided dialogue modeling pipeline.

\textbf{Metric design considerations:} We chose Coefficient of Variation ($CoV$) over standard deviation to represent variability since normalizing by the mean allows for comparison of variability across dialogue modeling components such as DST and NLG, as well as between two models with differing absolute performance.

For the standard deviation used in the numerator of $CoV$, we employ the sample standard deviation because we view the $K$ variants as a sample of the total population of possible ways a schema could be written. Using the sample standard deviation instead of the population standard deviation reduces bias of the estimate of the true variability.

Finally, by computing the average turn-level $CoV$ instead of computing $CoV$ on the dataset's top-line performance, we increase the metric's sensitivity to changes in prediction stability. Designing $SS$ as the average turn-level $CoV$ also provides us with a sense of how much a model's predictions can be expected to fluctuate at each given turn depending on how the schema is written.

\subsection{General Evaluation on SGD-X}
\label{sec:traineval}
In order to evaluate on SGD-X, we propose the following steps:

\begin{enumerate}
\item Train models on the original SGD train set schemas
\item Make predictions on the evaluation set using the 5 SGD-X schemas
\item Finally, measure performance on two classes of metrics:
\begin{enumerate}
    \item An average of standard performance metrics over the 5 variants
    \item Schema sensitivity metrics corresponding to the standard performance metrics
\end{enumerate}
\end{enumerate}

Using this training and evaluation setup best measures a model's ability to generalize to schemas written by a diverse set of authors.

\subsection{Dialogue State Tracking on SGD-X}
Because schema-guided dialogue state tracking (DST) is relatively well-studied, we apply the recommendations from section \ref{sec:traineval} and outline the evaluation procedure on SGD-X. We propose scoring DST models on 2 metrics: Average Joint Goal Accuracy ($JGA_{v_{1-5}}$) and Schema Sensitivity of JGA ($SS_{JGA}$).

We first compute model predictions across each of the $|T|$ dialogue turns in the eval set $|K|$ times - once for each of the schema variants - for a total of $|T| * |K|$ predictions.

We compute the average turn-level JGA as follows:

\begin{equation} \label{eq4}
JGA_{v_{1-5}} = \frac{\sum\limits_{t=1}^{T}\sum\limits_{k=1}^{K} JGA(p^k_t, g_t)}{|T|* |K|}
\end{equation}

Next, schema sensitivity $SS$ of the JGA is calculated following Equation (\ref{eq1}).

Note: in this evaluation, we only use predictions on the SGD-X variant schemas and not the original SGD schemas to avoid models ``cheating" by overfitting on the original schemas' writing styles.

We expect that $JGA_{v_{1-5}}$ will typically be the primary metric and $SS_{JGA}$ an auxiliary metric. The precise tradeoff between the two metrics when evaluating candidate models will depend on the context in which the model will be applied (e.g. how do we value higher accuracy vs. prediction consistency?). In the next section, we apply this evaluation on two DST models.

\begin{figure}[t]
\centering
\includegraphics[width=0.98\columnwidth]{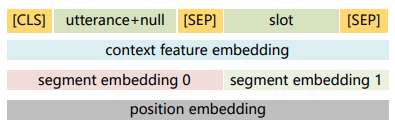}
\caption{Input to one of the four sub-models of SGP-DST responsible for free-form slot value prediction. The last 2 dialogue utterances, a ``null" token, and the slot description are concatenated (green), and the context feature takes on a value based on the slot's presence in the dialogue prior to this turn. After encoding, a slot value is predicted by selecting a span from the user utterance. Figure borrowed from \citet{ruan2020fine}.}
\label{fig:sgp-dst}
\end{figure}

\begin{figure*}[t]
\centering
\includegraphics[width=0.95\textwidth]{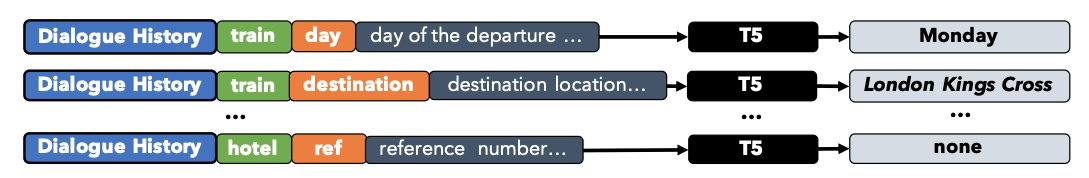}
\caption{Example inputs and outputs for fine-tuning the T5DST model. The model is run once for each slot. The dialogue history (blue), service (green), slot name (orange), and slot description (dark gray) are input to the model, and the predicted value is decoded. Figure borrowed from \citet{lee2021dialogue}.}
\label{fig:t5dst}
\end{figure*}

\section{Experiments}
Given schema-guided modeling for DST is relatively well studied, we use SGD-X to conduct two classes of robustness experiments:
\begin{enumerate}
    \item We train models on \textit{original SGD} and evaluate on \textit{SGD-X}
    \item We experiment with data augmentation techniques to improve performance on SGD-X
\end{enumerate}

\noindent We use the following models for our experiments: 

\begin{itemize}
    \item \noindent\textbf{SGP-DST}\footnote{While the authors of SGP-DST report 72.2\% JGA on the original SGD test set, we were only able to reproduce 60.5\% JGA when training with the recommended hyperparameters.} \cite{ruan2020fine} - the highest-performing model with publicly available code, at the time of writing. 4 sub-models are trained from independent BERT-Base encoders, each specializing in a sub-task. Each one takes the dialogue and relevant schema element names/descriptions as input and makes predictions, which are then combined across the 4 models using rules. Figure \ref{fig:sgp-dst} illustrates one sub-model.
    \item \noindent \textbf{T5DST} \citep{lee2021dialogue} - a generative model trained by fine-tuning T5-Base \citep{raffel2020exploring} to predict slot values given the dialogue context, service, slot name, and slot description, which achieves SOTA results on MultiWOZ 2.2. Figure \ref{fig:t5dst} depicts the model input and output.
\end{itemize}

\subsection{Train on SGD, Evaluate on SGD-X}
\label{sec:train-sgd-eval-sgdx}
We trained both models on the original SGD training set with the settings that produce their reported results, and then evaluated them on the SGD-X test sets. More training details in the Appendix.

\begin{table}[t]
\centering
\setlength{\tabcolsep}{3pt}

\begin{tabular}{|l|l|c|c|c|c|}
\hline
Model & Eval set & $JGA_{Orig}$ & $JGA_{v_{1-5}}$ & $Diff_{rel}$ & $SS_{JGA}$ \\\hline
\multirow{3}{*}{\notshortstack{SGP-\\DST}}  & all & 60.5 & 49.9 & \textbf{-17.6} & 51.9\\
&  seen & 80.1 & 60.7 & \textbf{-24.3} & 51.5\\
&  unseen & 54.0 & 46.3 & \textbf{-14.3} & 52.0\\
\hline
\multirow{3}{*}{\notshortstack{T5DST}} & all & 72.6 & 64.0 & \textbf{-11.9} & 40.4\\
& seen & 89.7 & 79.3 & \textbf{-11.6} & 31.9\\
& unseen & 66.9 & 58.9 & \textbf{-12.0} & 43.3\\
\hline
\end{tabular}

\caption{Evaluation of two top-performing DST models on the SGD-X test set. Both models experience substantial declines in performance when exposed to variant schemas.}
\label{tab:robustness-issue}
\end{table}

\begin{figure}[t]
\centering
\includegraphics[width=0.98\columnwidth]{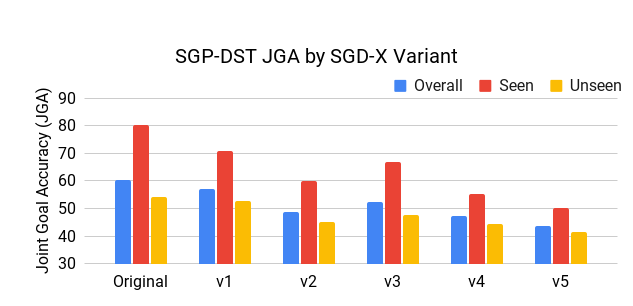}\qquad
\includegraphics[width=0.98\columnwidth]{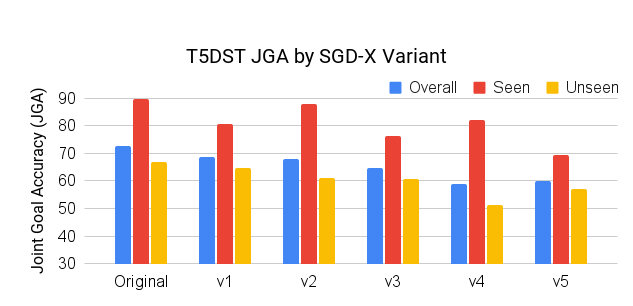}
\caption{JGA achieved by SGP-DST and T5DST respectively on the test set for the original SGD dataset and the five SGD-X variants. Both models fail to generalize well to variants of the original schemas.}
\label{fig:robustness-issue}
\end{figure}

\textbf{Results:} Table \ref{tab:robustness-issue} shows the summarized results and Figure \ref{fig:robustness-issue} displays JGA by variant. Both models see significant drops in joint goal accuracy, with SGP-DST and T5DST declining -$17.6\%$ and -$11.9\%$ respectively on average. For both models, the decline in JGA tends to increase in magnitude as the distance from the original schemas (reflected by the variant number) increases, with the two models dropping as much as -$28\%$ and -$19\%$ respectively for their worst variants. These results reveal that evaluating solely on the original SGD dataset overestimates the generalization capability of schema-guided DST models.

For SGP-DST, the JGA drop is much greater for seen services than unseen services. Recall that in this setup, ``seen" schemas at evaluation time are no longer linguistically identical to the schemas the models were trained on. The sharp decline suggests that SGP-DST likely overfit to the exact language used in seen schemas. Performance on unseen schemas also declines for both models, which we hypothesize is due to overfitting on the linguistic styles in the original SGD dataset, as mentioned in Section \ref{sec:intro}.

On schema sensitivity, T5DST scores almost $12$ points lower than SGP-DST in addition to achieving higher $JGA_{v_{1-5}}$, indicating it is superior to SGP-DST in both dimensions.

We observe that both models face robustness issues despite having powerful pre-trained language models as their base encoders, which have demonstrated immense success when applied to a variety of natural language tasks. We hypothesize that the models lose some of their generalization capabilities during the fine-tuning stage, a phenomenon also observed in other settings \citep{2020}.

\subsection{Schema Augmentation}

The results in Section \ref{sec:train-sgd-eval-sgdx} suggest both models overfit on the training schemas, reducing their ability to generalize to new linguistic styles. We experiment with back-translating schemas \citep{sennrich2016improving} to augment the training data \citep{hou2018sequence,yoo2019data} and study its impact on model robustness. In addition, to establish an approximate upper-bound for how much improvement paraphrasing-based schema augmentation can provide, we also evaluate the impact of augmenting the SGD-X crowdworker-collected paraphrases.

\textbf{Back-translation:} For each training schema, we back-translate its schema element names and descriptions three times using Google Translate to create three alternate schemas: one each for Mandarin, Korean, and Japanese - chosen for their relatively high difficulty and consequent diversity of back-translated paraphrases. The average normalized Levenshtein distance for names and BLEU score for descriptions between the originals and their back-translations are 0.14 and 34.1 respectively. Self-BLEU among back-translated variant schemas is 41.8. These metrics indicate a moderate degree of linguistic deviation from the original schemas and intra-variant diversity, though still much less than the SGD-X variants, which average 0.48 Levenshtein distance, 7.9 BLEU, and 4.5 self-BLEU. Examples pictured in Figure \ref{fig:schema-variations-bt}.

Once these variant schemas were created, new training examples were generated using the same dialogues as the original training set, but with schema inputs drawn from the variant schemas. When training on the augmented dataset, models encounter the same dialogue multiple times in a given epoch, where schema element names and descriptions differ for each version.

\begin{table}[t]
\centering
\setlength{\tabcolsep}{3pt}
\begin{tabular}{|l|l|c|c|}
\hline
Model & Aug method & $JGA_{v_{1-5}}$ & $SS_{JGA}$ \\\hline

\multirow{3}{*}{\notshortstack{SGP-\\DST}} & None & 49.9 & 51.9\\
& Backtrans & 54.1 (+8\%) & 43.1 (-17\%)\\
& Oracle & 66.2 (+33\%) & 22.5 (-57\%)\\\hline

\multirow{3}{*}{\notshortstack{T5DST}} & None & 64.0 & 40.4\\
& Backtrans & 70.8 (+11\%) & 34.0 (-16\%)\\
& Oracle & 73.3 (+15\%) & 24.6 (-39\%)\\

\hline
\end{tabular}
\caption{Results for schema augmentation methods on SGP-DST and T5DST models. Back-translation improves robustness for both models. Oracle augmentation, which involves augmenting SGD-X variant schemas, serves as an approximate upper bound for paraphrasing-based augmentation methods.}
\label{tab:bt-methods}
\end{table}

\textbf{SGD-X Crowdsourced Paraphrases (Oracle):} During crowdsourcing, we collected paraphrases for all 45 schemas across train, dev, and test sets. Similarly to the back-translation experiment, for this experiment we use the crowdsourced $v_1$ through $v_5$ training set schemas to augment the training data.  Note that this approach should be seen as an oracle for paraphrasing-based schema augmentation since this involves collecting roughly 5K human paraphrases for schema element names/descriptions. Furthermore, for a given variant ${v_i}$, the schema is the same for a service across train and eval sets. This means that models have already been exposed to the exact language used in seen schemas during training, giving them an unfair advantage on those services during evaluation.

\begin{figure}[t]
\centering
\includegraphics[width=0.98\columnwidth]{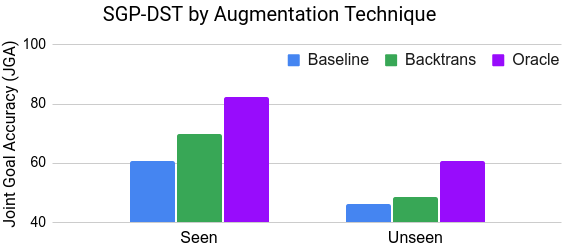}
\qquad
\includegraphics[width=0.98\columnwidth]{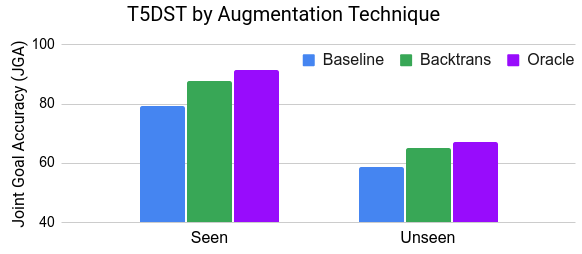}
\caption{$JGA_{v_{1-5}}$ for the SGP-DST and T5DST models with different schema augmentation methods, split by seen and unseen services. Back-translation improves performance across the board.} 
\label{fig:bt-methods}
\end{figure}

\begin{table*}[t]\centering
  \begin{tabular}{p{0.1\linewidth}|p{0.22\linewidth}|p{0.3\linewidth}|p{0.13\linewidth}}
    Service & Dialogue & Slot Name and Description & Predicted Value \\ \hline
    
    \multirow{2}{2cm}{Weather (seen)}
    & \multirow{2}{4cm}{USER: What will the weather in Portland be on the 14th?}
    & O: city - Name of the city

    & \textbf{Portland} \\ \cline{3-4}
    & & $v_1$: city name - Name of place & \textbf{\textit{None}}
    
    \\ 
    \hline
    
    \multirow{3}{2cm}{Payment (unseen)}
    & \multirow{3}{4cm}{USER: I need to make a payment from my visa.
    }
    & O: payment method - The source of money used for making the payment
    & \textbf{credit card} \\ \cline{3-4}
    & & $v_5$: money withdrawal source - What is being used to pay, either app balance or debit/credit card & \textbf{app balance} \\
    \hline
  \end{tabular}
  \caption{Examples where T5DST fails to predict slots correctly when given SGD-X variant schemas. O represents the original, and $v_i$ represents the i-th SGD-X schema.}
\label{tab:err-examples}
\end{table*}

\textbf{Results:} We train the SGP-DST and T5DST models using the two aforementioned schema augmentation approaches and evaluate on the SGD-X benchmark (without augmentation). The results are summarized in Table \ref{tab:bt-methods} and Figure \ref{fig:bt-methods}. 

Training with back-translated schemas improves the robustness of both models. Accuracy on SGD-X increases by +8\% for SGP-DST and +11\% for T5DST (relative), and it decreases schema sensitivity -17\% and -16\%, respectively. The improvement is considerable for unseen as well as seen schemas, suggesting that training with diverse schemas improves model generalization. This result is consistent with \citet{wei2021finetuned}, which hypothesizes that increasing diversity of training data improves performance on unseen tasks. The oracle method further improves joint goal accuracy and schema sensitivity beyond back-translation - a useful reference for how much paraphrasing-based schema augmentation may improve performance. 

Although the models trained with back-translation do not achieve parity with performance on the original SGD test set (54.1\% vs. 60.1\% JGA for SGP-DST, 70.8\% vs. 72.6\% for T5DST), much of the decline is recouped. Not only is this technique effective, but it is easy to implement, model agnostic, and requires requires no changes to modeling code. 

Given that back-translating schemas with Mandarin, Korean, and Japanese already produces a relatively high BLEU score of 34.1 despite being tough to translate, we hypothesize that incorporating additional back-translated schemas from other languages would not greatly increase the diversity of linguistic styles. As a result, we believe that simply scaling to more languages would yield limited improvements in performance. One alternative to further increase linguistic diversity would be to introduce sampling when decoding for back-translation.

\textbf{Other augmentation methods:} Besides back-translation, we also experimented with augmenting corrupted versions of schemas, where we randomly replaced words and perturbed word order. However, we did not see improvements over the non-augmented models, which we hypothesize is due to a mismatch between the corrupted training schemas and real test schemas. Besides augmenting schemas, augmenting dialogues has shown promise in other settings and could also improve robustness \citep{ma2019end, noroozi2020fast}.

\section{Analysis}

To gain better intuition of model robustness issues, we inspect cases where T5DST predicts incorrectly when given variant schemas. We also analyze T5DST's performance broken down by service. All analysis is done on T5DST trained only on the original SGD schemas.

\subsection{Visually Inspecting Errors}

We visually inspected examples where T5DST fails to predict slots correctly when provided with variant schemas. Many errors arise from failing to predict slots as active. For example, in the Weather dialogue in Table \ref{tab:err-examples}, the model correctly predicts ``city = Portland" when given the original schema but mis-predicts ``city name = None" for the $v_1$ variant. In these cases, the model may not understand the slot name and description well, possibly leading it believe the slot is irrelevant for the current dialogue.

We also observe cases where the model correctly predicts a categorical slot as active but predicts an incorrect value. For example, in the Payment dialogue in Table \ref{tab:err-examples}, the model predicts that the slot for ``money withdrawal source" is ``app balance" instead of ``credit card" when given the $v_5$ schema. One hypothesis is that the word ``withdrawal" in the name ``money withdrawal source" biases the model to decode ``balance" over ``credit card", since ``balance" and ``withdrawal" are words present in the Banks schema seen at training time.

While SGD's original schemas and SGD-X variant schemas are semantically similar from a human's perspective, these slight perturbations have an outsized impact on model performance, highlighting the degree to which models overfit on the writing styles of schemas.

\begin{figure}[t]
\centering
\includegraphics[width=1.0\columnwidth]{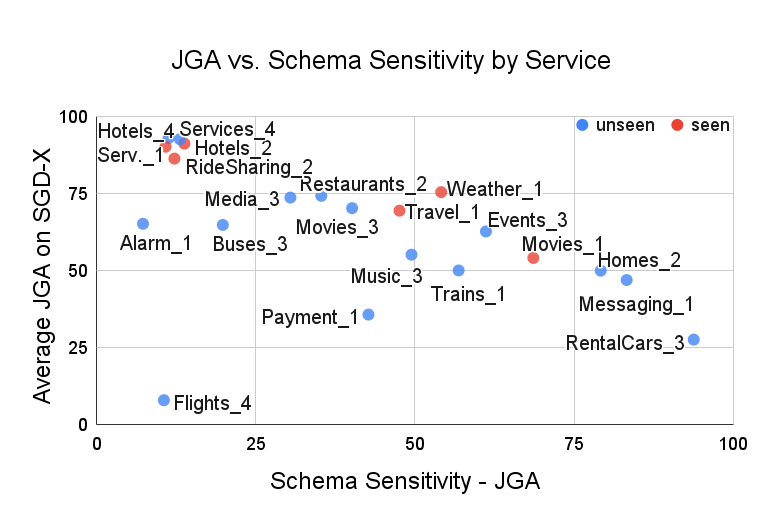}
\caption{A plot of Average Joint Goal Accuracy and Schema Sensitivity on the test set for the T5DST model trained only on original SGD. Each point represents one service. The model tends to be less sensitive to schema variations for services it predicts more accurately.}
\label{fig:jga-vs-ss}
\end{figure}

\subsection{Service-level Results}
In order to dissect model performance further, we plot the Average JGA ($JGA_{v_{1-5}}$) and Schema Sensitivity to JGA ($SS_{JGA}$) by service, shown in Figure \ref{fig:jga-vs-ss}. We observe that higher $JGA_{v_{1-5}}$ tends to correspond to lower $SS_{JGA}$. This suggests that higher accuracy prediction stability come hand in hand, for both seen and unseen services.

Given how $SS_{JGA}$ is defined, for a given service, a model could predict the dialogue state inaccurately yet also achieve a desirably low schema sensitivity. However, our results suggest that this is atypical, with \texttt{Flights\_4} being one of the exceptions to this pattern. We hypothesize that \texttt{Flights\_4} breaks this pattern because it is exceptionally challenging to predict its state, leading the model to make uniformly poor predictions regardless of which schema variant it is given.

\section{Related Work} 
Schema-guided modeling aims to build task-oriented dialogue systems that can generalize easily to new verticals using very little extra information, including for slot filling \citep{bapna2017towards, shah2019robust, liu2020robustness} and dialogue state tracking \citep{li2021zero,campagna2020zero,kumar2020ma} among other tasks. More recent work has adopted the schema-guided paradigm \citep{ma2019end, li2020sppd, zhang2021sgd} and even extended the paradigm in functionality \citep{mosig2020star, mehri2021schema}.

Model robustness is an active area of NLP research \citep{goel2021robustness} and has many interpretations, such as to noise \citep{belinkov2018synthetic}, distribution shift \citep{hendrycks2020many} and adversarial input \citep{jia2017adversarial}.

As they are inherently public-facing in nature, the robustness of dialogue systems to harmful inputs \citep{dinan2019build,cheng2019evaluating} and input noise \citep{einolghozati2019improving, liu2020robustness}, such as ASR error, misspellings, and user input paraphrasing have been explored. However, robustness to API schemas for schema-guided dialogue systems remains relatively unexplored.

\citet{lin2021leveraging} and \citet{cao2021comparative} both investigate natural language description styles for zero/few-shot dialogue state tracking. The former experiments with homogeneously training and evaluating on different description styles, unlike our work. The latter performs heterogeneous evaluation of template-based description styles (e.g. rephrasing slot name as a question, using the original description). Models are also evaluated against paraphrased descriptions created via back-translation but only decline slightly in performance.

\section{Conclusion}

In this work, we present SGD-X, a benchmark dataset for evaluating the robustness of schema-guided models to schema writing styles. To evaluate robustness, we propose training models on SGD, predicting on SGD-X, and finally measuring standard performance metrics alongside a novel \textit{schema sensitivity} metric that quantifies the stability of model predictions across variants.

Applying this to two of the highest-performing schema-guided DST models, we discover that both perform substantially worse on SGD-X than SGD, suggesting that evaluating solely on SGD overestimates models’ ability to generalize to real-world schemas. It’s noteworthy that we witness this decline on models based on T5 and BERT - two popular large language models in research and production. We further demonstrate that back-translating schemas for training data augmentation is an effective, model-agnostic technique for recovering some of this decline while simultaneously reducing schema sensitivity.

We note that the weaknesses of evaluating only on the original SGD dataset uncovered in this work also apply to the leave-one-domain-out zero-shot evaluation on the popular MultiWOZ dataset. Also, while dialogue state tracking is the focal point of this work, SGD-X is applicable to evaluating the robustness of other schema-guided dialogue components (e.g. policy, NLG). We hope that releasing this paper and benchmark motivates further research in the area of schema robustness.

\section{Ethical Impact}

\textbf{Crowdsourcing details:} We hired 400+ Amazon Mechanical Turk crowdworkers from the U.S. and paid USD \$1-2 per task, where each task consisted of paraphrasing either names or descriptions for every element in a single schema. The median submission time was 3 minutes, which equates to US\$20-40/hr. In total, we spent $\sim$\$2000 on data collection.

\bibliography{aaai22}

\appendix

\section{Training Details}
\label{sec:training_details}

\begin{figure*}[b]
\centering
\includegraphics[width=0.98\textwidth]{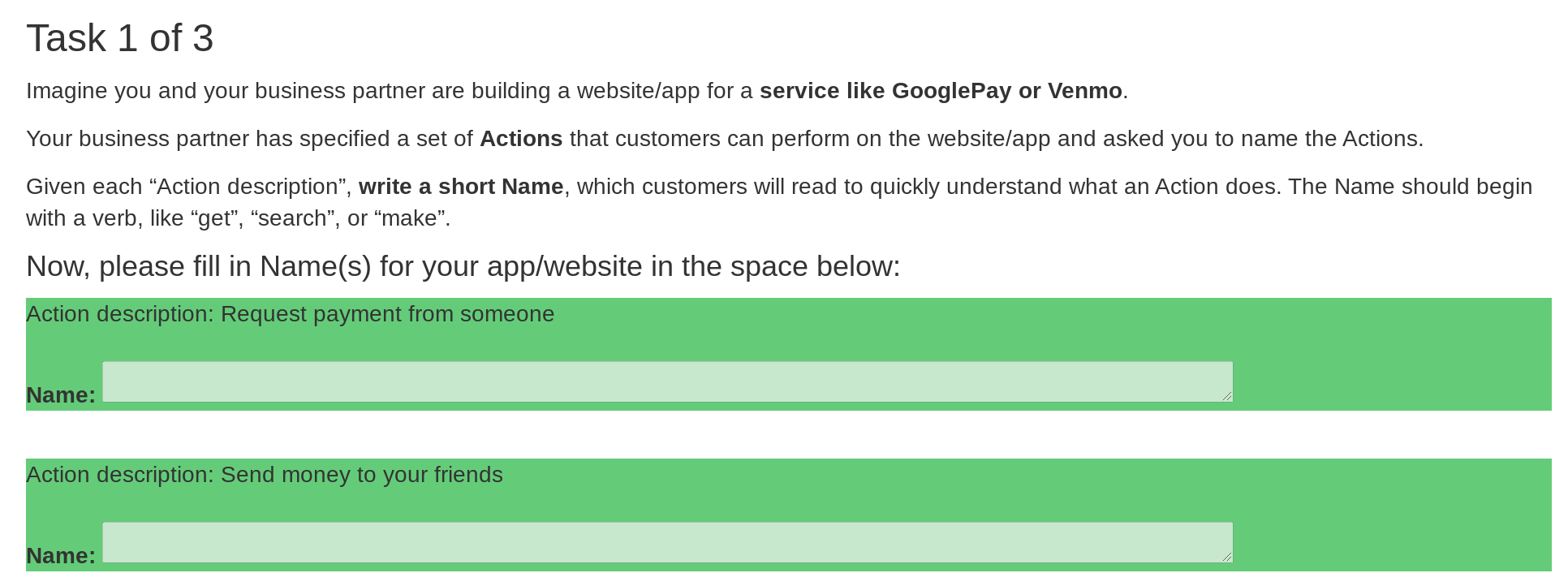}
\caption{An example of the task shown to the crowdworker to obtain an intent name paraphrases.}
\label{fig:endor-intent-name}
\end{figure*}

\begin{figure*}[b]
\centering
\includegraphics[width=0.98\textwidth]{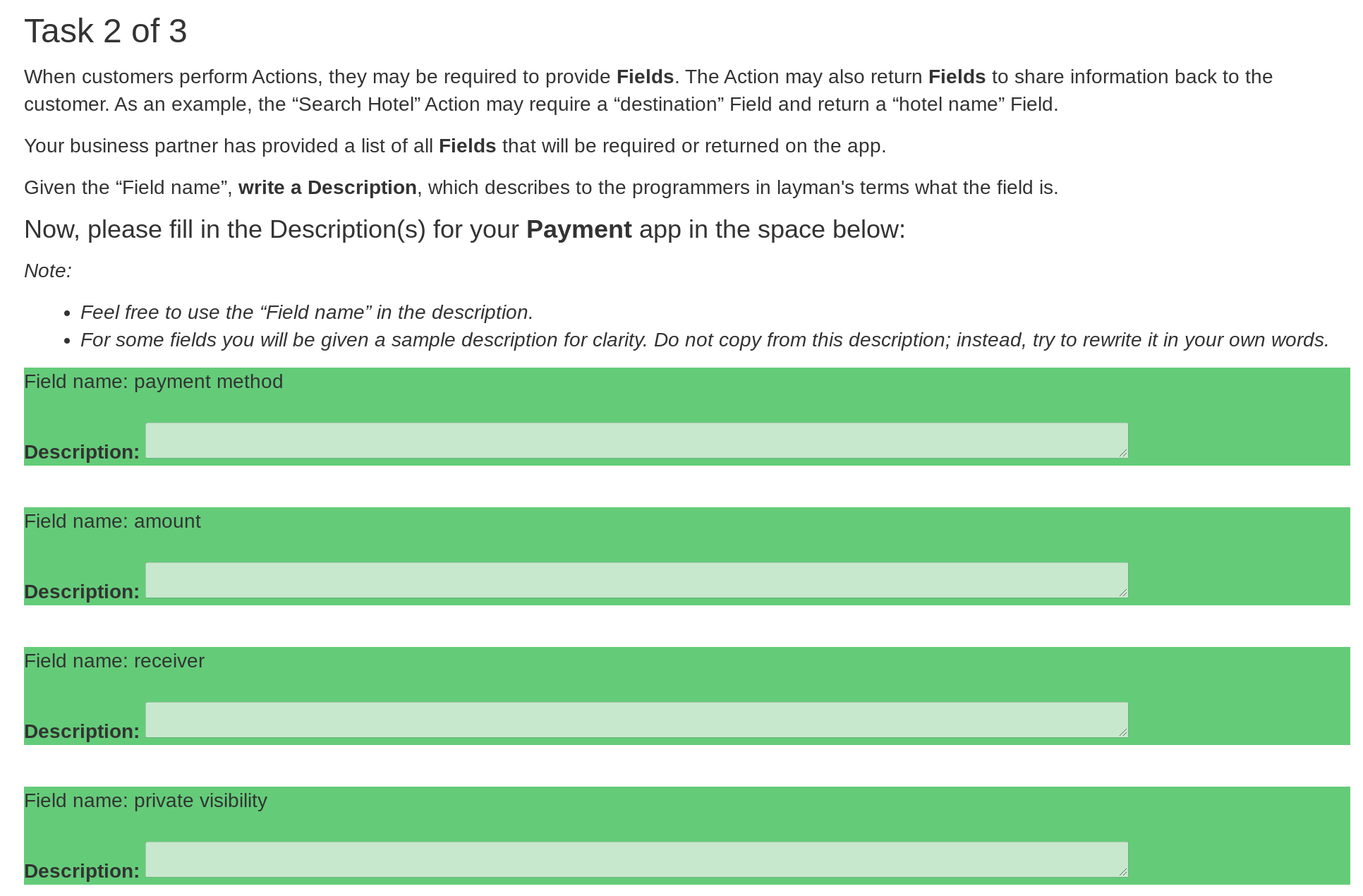}
\caption{An example of the task shown to the crowdworker to obtain slot description paraphrases.}
\label{fig:endor-slot-desc}
\end{figure*}

For training the SGP-DST model \citep{ruan2020fine}, we used the BertAdam optimizer in PyTorch \citep{kingma2014adam, paszke2019pytorch} with learning rate 2e-5, warmup proportion 0.1 (except the \texttt{copy\_slot} BERT encoder within this model, which had a warmup proportion of 0.7), training batch size 32, and default settings for other parameters of the BertAdam optimizer. Each of the BERT encoders in the model was trained for 1 epoch (generally converging earlier), except the \texttt{combine} BERT encoder, which was trained for 3 epochs for the baseline SGP-DST model, and 1 epoch for all experiments with augmented training data. Depending on the training data size, training took between 24-36 hours on a single NVIDIA Tesla V100-SXM2 GPU core.

\begin{figure*}[h]
\centering
\includegraphics[width=0.98\textwidth]{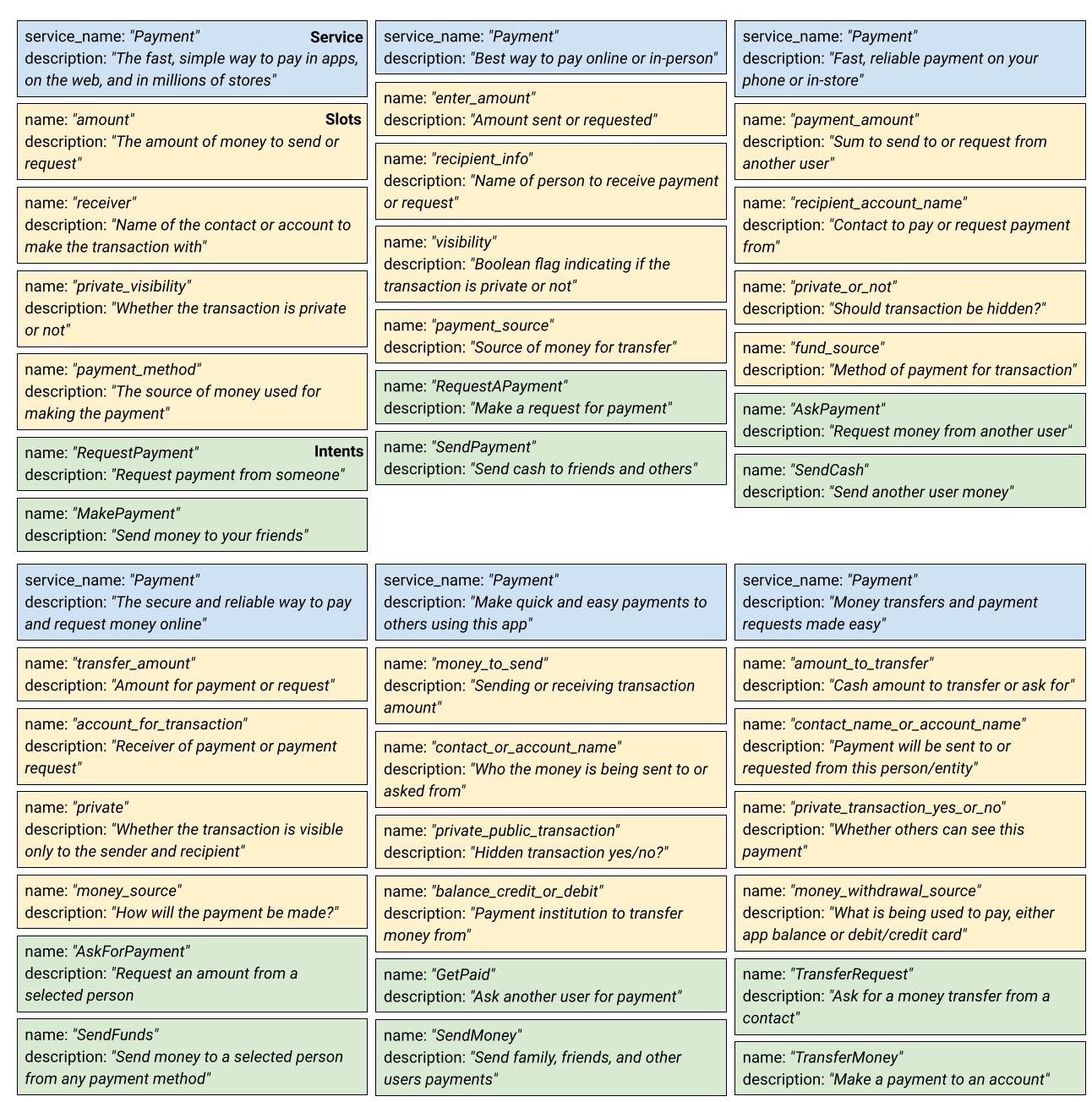}
\caption{An original schema from the SGD dataset and its five variant schemas generated from crowdworker paraphrases, in order (original, $v1$, $v2 \dots v5$ from left-to-right, top-to-bottom).}
\label{fig:schema-variations-cw}
\end{figure*}

\begin{figure*}[h]
\centering
\includegraphics[width=0.98\textwidth]{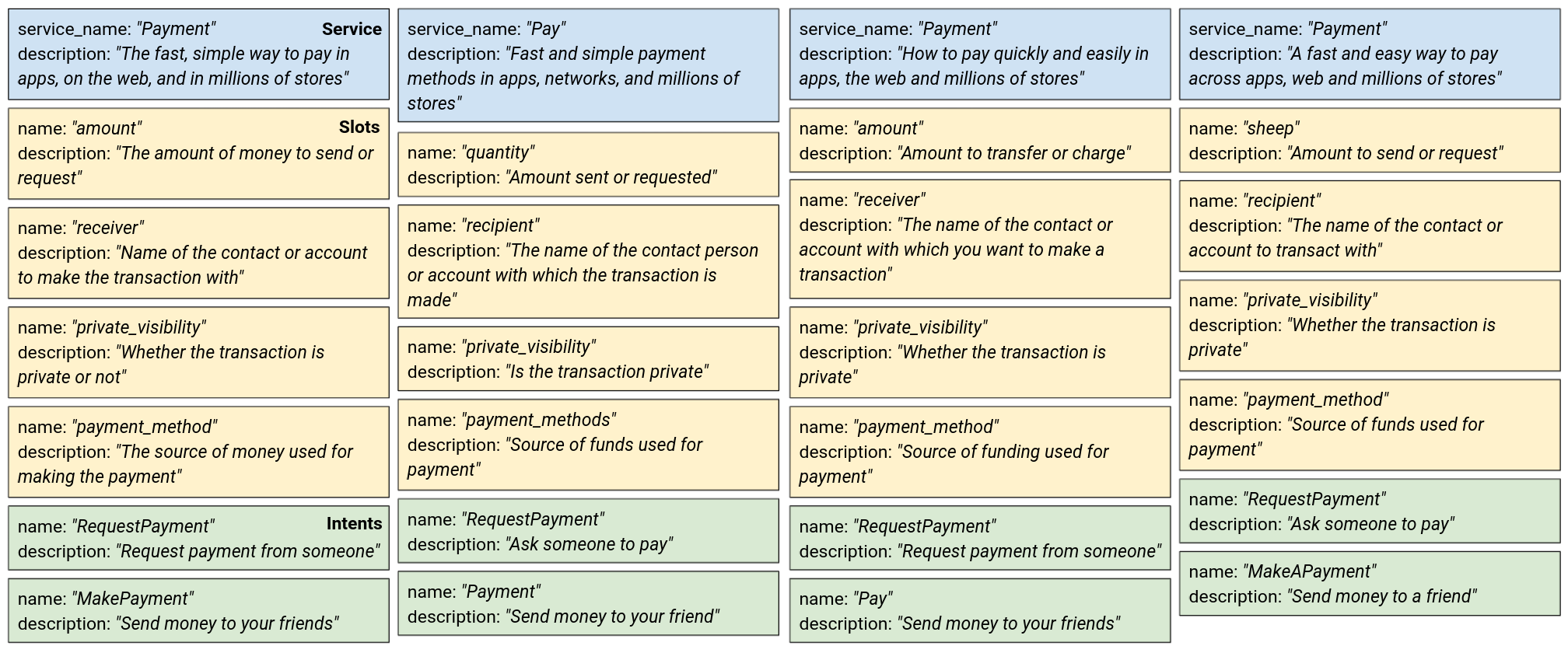}
\caption{An original schema (left) from the SGD dataset and its three variant schemas generated through backtranslation (Mandarin Chinese, Japanese, and Korean - left to right).}
\label{fig:schema-variations-bt}
\end{figure*}

The T5DST model, in contrast, was trained with 32 TPU v3 chips \citep{jouppi2017datacenter}. For fine-tuning, we use batch size 16 and use constant learning rate of 1e-4 across all experiments for 50,000 finetuning steps. The input and output sequence lengths are 512 and 256 tokens, respectively.

\section{Crowdworker Task}
Figures \ref{fig:endor-intent-name} and \ref{fig:endor-slot-desc} show snippets of the crowdworker tasks for collecting intent name paraphrases and slot description paraphrases respectively. Note that the original values for both were omitted to encourage diversity in the crowdworker responses.

\section{Example Schemas}

Figure \ref{fig:schema-variations-cw} shows an example of a schema from the dataset for a digital wallet service along with its five SGD-X variant schemas. The names and descriptions are increasingly distant from the original, and there exists sufficient variation between the variant schemas themselves. Figure \ref{fig:schema-variations-bt} shows the original schema along with its three back-translated variants. These back-translated schemas present far less diversity in the paraphrases compared to the human-paraphrased variant schemas and introduce errors as well (e.g., \textit{"amount"} $\rightarrow$ \textit{"sheep"} in the Korean back-translated schema).

\end{document}